\documentclass[10pt,twocolumn]{article}

\usepackage[a4paper,margin=0.75in]{geometry}
\usepackage[T1]{fontenc}
\usepackage[utf8]{inputenc}
\usepackage{lmodern}

\usepackage{amsmath, amssymb}
\usepackage{graphicx}
\usepackage{booktabs}
\usepackage{tabularx}
\usepackage{cite}
\usepackage{url}
\usepackage[hidelinks]{hyperref}
\usepackage{authblk}
\usepackage{orcidlink}
\usepackage{balance}

\usepackage{fvextra}

\usepackage{caption}
\title{LLM-Based vs. Lexicon-Based Sentiment Signals for Tail-Risk Detection in Meme Stocks}

\author[1]{Paul Kilian}
\author[1]{Markus Kleffmann\orcidlink{0009-0004-4175-552X}}
\affil[1]{
    IU International University of Applied Sciences, Erfurt, Germany\\
    \href{mailto:paul.kilian@iu-study.org}{paul.kilian@iu-study.org}, %
    \href{mailto:markus.kleffmann@iu.org}{markus.kleffmann@iu.org}
}

\date{}

\begin{document}

\maketitle

\begin{abstract}
This paper presents an empirical comparison of lexicon-based and Large Language Model (LLM)-based sentiment analysis for extracting market-relevant signals from social media discourse in highly volatile equity markets. Using Reddit data from r/WallStreetBets and focusing on meme stocks (GME, AMC, NOK), we construct time-aligned sentiment indicators and evaluate their relationship with market returns, with particular attention to extreme positive return events in the upper tail of the return distribution. The LLM-based approach generates multidimensional sentiment representations capturing emotional polarity, bullishness, sarcasm likelihood, and topical relevance, whereas the baseline relies on the VADER lexicon-based model. We evaluate both approaches using lead/lag correlation analysis, OLS regression, ROC-AUC-based directional classification, and a quantile-based early-warning framework. The results indicate that LLM-derived indicators provide a richer multidimensional representation and exhibit stronger asset-specific statistical structure than the lexicon-based baseline. However, their relationship with market movements remains heterogeneous across assets, suggesting that increased linguistic expressiveness does not necessarily translate into stable forecasting performance in retail-driven volatility regimes.
\end{abstract}

\vspace{0.5em}


\vspace{0.5em}

\section{Introduction}

Social media platforms have become an increasingly important factor in financial market dynamics, particularly during periods of elevated volatility and retail-driven trading activity \cite{barber2022, andreev2022}. Empirical research suggests that investor attention and sentiment extracted from textual data can influence short-term price formation and return dynamics \cite{tetlock2007, bollen2011}.

This phenomenon is especially pronounced in so-called meme stocks, where coordinated retail attention has been associated with abrupt and extreme price movements \cite{barber2022, andreev2022}.

Sentiment analysis is widely used to extract structured signals from unstructured textual data in financial contexts \cite{tetlock2007, bollen2011, kearney2014}. Traditional lexicon-based approaches such as VADER rely on predefined dictionaries and heuristic rules to infer sentiment polarity \cite{hutto2014}. While these methods are computationally efficient and interpretable, they are limited in their ability to capture contextual phenomena such as sarcasm, slang, and domain-specific language, which are prevalent in social media discourse and are particularly challenging to capture with rule-based approaches alone \cite{hutto2014, agarwal2011}.

Recent advances in Natural Language Processing, particularly transformer-based Large Language Models (LLMs), address some of these limitations by incorporating contextual and semantic dependencies in text representation \cite{vaswani2017, devlin2019}. However, long-context processing and robust information aggregation remain non-trivial challenges in such models \cite{liu2024}.

Recent studies suggest that transformer-based language models and LLMs can improve sentiment classification and textual analysis in financial domains, although empirical evidence regarding their ability to generate consistently informative or predictive market signals remains mixed \cite{chen2024, araci2019, lopezlira2026}. In particular, it is still unclear whether improved linguistic representations translate into stable predictive advantages in noisy, retail-driven financial environments.

This study investigates this question by empirically comparing a lexicon-based sentiment model (VADER) with a Large Language Model-based sentiment framework applied to Reddit discourse from the r/WallStreetBets community. The analysis focuses on three highly volatile equities (GME, AMC, NOK) and evaluates the relationship between extracted sentiment signals and subsequent market returns in tail-risk regimes. In this study, tail-risk events are operationalized specifically as extreme positive returns in the upper tail of the asset-specific return distribution.

The study is guided by the following research questions:
\begin{itemize}
    \item \textbf{RQ1:} How do lexicon-based and LLM-based sentiment models differ in the dimensionality and structure of the indicators extracted from financial social media discourse?
    
    \item \textbf{RQ2:} To what extent do VADER- and LLM-derived indicators exhibit predictive or reactive relationships with stock returns, and how do they compare in short-term directional classification?
    
    \item \textbf{RQ3:} To what extent can a composite LLM-based discourse indicator provide early-warning information for extreme positive return events?
\end{itemize}

The remainder of this paper is structured as follows. Section~\ref{sec:related_work} reviews related work on sentiment analysis in financial contexts and the use of language models for textual market analysis. Section~\ref{sec:methodology} describes the data sources and the methodological framework, including sentiment extraction, aggregation, and evaluation design. Section~\ref{sec:results} presents the empirical results of the comparative analysis between lexicon-based and LLM-based approaches. Section~\ref{sec:discussion} discusses the findings and their implications for financial sentiment modeling in retail-driven markets. Section~\ref{sec:conclusion} concludes the paper and outlines directions for future research.

\section{Related Work}
\label{sec:related_work}

The relationship between textual sentiment and financial market dynamics has been widely studied in behavioral finance and computational economics. Early work provides evidence that textual measures of investor sentiment and attention are associated with asset prices and short-term return dynamics \cite{tetlock2007, kearney2014, bollen2011}.

A central line of research focuses on lexicon-based sentiment models, which rely on predefined dictionaries and rule-based heuristics to estimate sentiment polarity. Among these approaches, VADER has become a widely used baseline for sentiment analysis in social media contexts due to its efficiency and interpretability \cite{hutto2014}. However, lexicon-based methods are inherently limited in their ability to capture contextual ambiguity, informal language, and domain-specific expressions common in online discourse \cite{hutto2014, agarwal2011}. These limitations are particularly relevant in online financial communities such as r/WallStreetBets, whose discourse exhibits community-specific linguistic features and dominant phrases, while sarcasm and meme-driven expression pose additional challenges for sentiment classification in social media contexts \cite{agrawal2022, kader2022, hayakawa2024}.

To address such contextual limitations, machine learning-based approaches have increasingly been applied to sentiment classification. Transformer-based models such as BERT improve contextual language understanding through bidirectional attention mechanisms \cite{devlin2019}. In financial applications, domain-adapted variants such as FinBERT demonstrate improved performance in sentiment classification tasks compared to general-purpose models \cite{araci2019}.

More recently, Large Language Models (LLMs) have been explored for financial sentiment analysis and market-related text classification. These models can capture richer contextual and semantic information than traditional lexicon-based approaches, but evidence regarding their ability to generate consistently informative or predictive market signals remains mixed \cite{chen2024, lopezlira2026}.

A particularly relevant application domain is the analysis of retail-driven trading communities such as r/WallStreetBets, where collective attention and coordinated discourse have been associated with extreme price movements and increased market volatility \cite{barber2022, andreev2022}. Much of the broader financial sentiment literature relies on aggregated sentiment measures or lexicon-based textual indicators for modeling financial text data \cite{tetlock2007, bollen2011, kearney2014}. While recent work has begun to explore transformer-based and LLM-based methods, systematic comparisons of these approaches against traditional sentiment models remain comparatively limited in financial market settings \cite{chen2024, lopezlira2026}.

Against this background, this paper provides a direct empirical comparison between a lexicon-based sentiment model (VADER) and a Large Language Model-based multidimensional sentiment framework in the context of social media-driven equity markets and tail-risk events.

\section{Data and Methodology}
\label{sec:methodology}

\subsection{Research Design}

This study follows a quantitative and comparative research design. The objective is to evaluate whether LLM-based sentiment representations provide richer and potentially more informative market signals than a lexicon-based baseline in highly volatile, retail-driven equity markets. The empirical analysis focuses on three meme stocks: GameStop (GME), AMC Entertainment (AMC), and Nokia (NOK). These equities were selected because they exhibited pronounced volatility and unusually high levels of social media attention during the meme-stock period around early 2021.

The analysis is conducted at daily frequency. Reddit entries, including both posts and comments, are transformed into sentiment-based indicators and aligned with historical stock market data. The resulting time series are then evaluated using correlation-based, regression-based, and classification-based methods.

\subsection{Data Sources}

The textual dataset consists of Reddit posts and comments from r/WallStreetBets. The observation window covers the period from November 2020 to May 2022. This period captures both the build-up and aftermath of the meme-stock events in early 2021.

The data were obtained from two complementary sources. First, the Reddit WallStreetBets Posts dataset published on Kaggle was used as a structured source of Reddit submissions \cite{preda2021}. Kaggle is an online data science platform that provides public datasets and computational notebooks for machine learning and data analysis projects \cite{kaggleDatasets}. Since the Kaggle dataset used in this study was not pre-filtered by individual stock, posts were selected using ticker-based filtering for GME, AMC, and NOK.

Second, asset-specific JSON archives from a Figshare repository were integrated to include a broader set of posts and comments \cite{longo2023}. Figshare is an online research repository infrastructure for sharing, publishing, and citing research outputs and datasets \cite{figshareAbout}. Historical daily closing prices were obtained using the \texttt{yfinance} Python package \cite{aroussi2026} and aligned with the Reddit data on a daily basis.

Table~\ref{tab:data_overview} summarizes the resulting dataset after ticker-based filtering and source integration. In the table, posts refer to original Reddit submissions that initiate a discussion thread, whereas comments refer to replies posted within such threads, either in response to the original post or to other comments. The strong imbalance between posts and comments reflects the different structure of the underlying data sources: the Kaggle dataset contributes Reddit submissions, while the Figshare archives primarily provide thread-level comment data.

\begin{table}[htbp]
\centering
\caption{Overview of Reddit Dataset by Asset and Source}
\label{tab:data_overview}
\begin{tabular}{llrrr}
\toprule
Asset & Source & Posts & Comments & Total \\
\midrule
GME & Kaggle & 13,585 & 0 & 13,585 \\
GME & Figshare & 8 & 489,755 & 489,763 \\
AMC & Kaggle & 5,379 & 0 & 5,379 \\
AMC & Figshare & 2 & 110,702 & 110,704 \\
NOK & Kaggle & 1,942 & 0 & 1,942 \\
NOK & Figshare & 2 & 1,985 & 1,987 \\
\midrule
Total & -- & 20,918 & 602,442 & 623,360 \\
\bottomrule
\end{tabular}
\end{table}

\subsection{Preprocessing}

The heterogeneous Reddit data sources were transformed into a unified analysis schema. For each entry, title and body fields were merged into a single raw text variable, timestamps were parsed into a common datetime format, and metadata such as Reddit identifier, score, source, and post type were retained. The Reddit identifier was used as a persistent primary key throughout the pipeline to ensure that sentiment outputs could be deterministically linked back to the original text entries.

Ticker-based filtering was applied where necessary, particularly to the Kaggle dataset, to assign entries to GME, AMC, and NOK. Asset-specific Figshare archives were merged into the corresponding asset-level datasets. Entries without interpretable textual content, empty placeholders, and bot-like automatically generated content were removed using heuristic filtering rules. Duplicate entries were removed based on the unique Reddit identifier.

Informal language features such as emojis, abbreviations, slang, and community-specific expressions were deliberately preserved. These elements are semantically relevant in r/WallStreetBets discourse and may contain information that is particularly important for contextual sentiment models.

\subsection{Lexicon-Based Baseline: VADER}

As a lexicon-based baseline, we use VADER, a rule-based sentiment model specifically designed for social media text \cite{hutto2014}. VADER assigns each text a compound sentiment score in the interval $[-1,1]$, where negative values indicate negative sentiment and positive values indicate positive sentiment.

For each asset and calendar day, two VADER-based sentiment indicators are computed. The first indicator is the unweighted daily average sentiment:

\begin{equation}
S^{VADER}_{t} = \frac{1}{n_t} \sum_{i=1}^{n_t} c_i,
\end{equation}

where $c_i$ denotes the VADER compound score of entry $i$ and $n_t$ denotes the number of Reddit entries for a given asset on day $t$.

The second indicator is a score-weighted sentiment index that incorporates the Reddit score:

\begin{equation}
S^{VADER,w}_{t} =
\frac{\sum_{i=1}^{n_t} c_i \cdot (score_i + 1)}
{\sum_{i=1}^{n_t} (score_i + 1)}.
\end{equation}

Here, $score_i$ denotes the raw Reddit score of entry $i$, i.e., the net community rating resulting from upvotes and downvotes. It is a metadata variable rather than a sentiment score and is used as an approximate proxy for community evaluation and engagement. The raw score is retained in the weighted aggregation. Consequently, entries with a score of $-1$ receive zero weight, while scores below $-1$ produce negative weights. The offset of one ensures that entries with a Reddit score of zero retain unit weight. The unweighted indicator captures the average sentiment expressed across all entries on a given day, whereas the weighted indicator incorporates both the direction and magnitude of community evaluation.

The resulting indicators serve as the lexicon-based benchmark against which the LLM-derived sentiment representations are evaluated.

\subsection{LLM-Based Sentiment Extraction}

For the LLM-based analysis, Reddit entries were classified using Gemini 2.5 Flash-Lite. The model was selected because it provides a scalable and cost-efficient option for large-scale batch inference over several hundred thousand short social media texts \cite{geminiModels2026, googleCloud2026}. Unlike the VADER baseline, which produces a single polarity score, the LLM-based approach generates a multidimensional representation of each Reddit entry.

The model configuration was held constant across all inference jobs. Default generation settings were used, and no fine-tuning, dynamic parameter adjustment, or prompt variation was applied. Inference was performed asynchronously through the official Gemini Batch API.

To control input length and ensure scalable inference, each Reddit entry was truncated to the first 400 characters before classification. Entries were processed in fixed-size batches of 50. During pipeline development, sample-based manual plausibility checks were used to assess whether community-specific slang, memes, and informal expressions were interpreted consistently. These checks were exploratory and did not constitute a manually annotated ground-truth evaluation.

For each entry, the model outputs four variables:

\begin{itemize}
    \item \textbf{Sentiment} ($s$): the general emotional polarity of the entry in the interval $[-1,1]$, where negative values indicate pessimistic or negative tone and positive values indicate optimistic or positive tone.
    
    \item \textbf{Bullishness} ($b$): the explicit or implicit market direction expressed in the entry in the interval $[-1,1]$, where negative values indicate bearish, selling, or short-oriented intent and positive values indicate bullish, buying, or holding-oriented intent.
    
    \item \textbf{Sarcasm} ($sc$): the estimated degree of sarcastic, ironic, or non-literal expression in the interval $[0,1]$.
    
    \item \textbf{Relevance} ($r$): the topical relevance of the entry to the respective stock or market event in the interval $[0,1]$.
\end{itemize}

This multidimensional structure separates emotional tone from market intent. This distinction is important in financial social media discourse because a text can express negative emotion while still conveying bullish trading intent, for example when users express anger toward short sellers while indicating willingness to hold or buy a stock.

The classification prompt enforced structured numerical output and preserved the original Reddit identifier for each entry. This ensured that model outputs could be deterministically linked back to the corresponding Reddit entry during data integration. The prompt contained no few-shot examples and required one minified JSON output for each input entry. The complete prompt is reproduced in Appendix~\ref{app:prompt}.

\subsection{LLM-Based Aggregation}

The LLM outputs are aggregated into daily indicators for each asset. Since the main objective is to capture market-relevant direction rather than general emotional tone alone, the primary aggregation is based on the bullishness variable $b_i$.

The first LLM-based indicator is the unweighted daily average of bullishness:

\begin{equation}
B^{LLM}_{t} = \frac{1}{n_t} \sum_{i=1}^{n_t} b_i,
\end{equation}

where $b_i$ denotes the LLM-derived bullishness score of entry $i$ and $n_t$ denotes the number of Reddit entries for a given asset on day $t$.

The second indicator is a score-weighted bullishness index:

\begin{equation}
B^{LLM,w}_{t} =
\frac{\sum_{i=1}^{n_t} b_i \cdot (score_i + 1)}
{\sum_{i=1}^{n_t} (score_i + 1)}.
\end{equation}

As in the VADER-based weighted index, $score_i$ denotes the raw Reddit score, and the same weighting convention is retained. Higher positive scores receive greater weight, while scores below $-1$ produce negative weights. The resulting indicator therefore incorporates both positive and negative community evaluation into the daily aggregation.

To address RQ3, we construct a composite Language-based Market Index (LMI) in addition to the one-dimensional bullishness indicators. The LMI integrates emotional polarity, market intent, topical relevance, sarcasm adjustment, and social visibility into a single entry-level signal. To ensure that the logarithmic transformation remains mathematically defined, negative Reddit scores are clipped at zero before transformation:

\begin{equation}
LMI_i = (s_i + b_i) \cdot r_i \cdot (1 - sc_i)
\cdot \ln\left(1 + \max(score_i,0)\right).
\end{equation}

Here, $s_i$ denotes emotional sentiment, $b_i$ denotes bullishness, $r_i$ denotes topical relevance, and $sc_i$ denotes the sarcasm score of entry $i$. The additive combination of sentiment and bullishness strengthens the signal when emotional tone and market intent point in the same direction, while partially offsetting inconsistent discourse signals. The relevance factor reduces the influence of entries with weak stock-specific content. The term $(1 - sc_i)$ down-weights sarcastic or non-literal expressions. The logarithmic transformation reduces the dominance of highly visible individual entries, while entries with non-positive Reddit scores receive a multiplicative weight of zero and therefore do not contribute to the LMI.

The daily LMI is computed as the average of all entry-level LMI values for a given asset and day:

\begin{equation}
LMI_t = \frac{1}{n_t} \sum_{i=1}^{n_t} LMI_i.
\end{equation}

Together, the pure bullishness index, the score-weighted bullishness index, and the LMI provide complementary perspectives on the LLM-derived discourse signal: the first captures average market intent, the second incorporates social amplification, and the third combines multiple semantic and social dimensions into a composite market indicator.

\subsection{Market Return Alignment}

Historical daily closing prices were aligned with the daily sentiment indicators for each asset. For the correlation, regression, and classification analyses, daily returns were computed from closing prices as:

\begin{equation}
R_t = \frac{P_t - P_{t-1}}{P_{t-1}},
\end{equation}

where $P_t$ denotes the closing price of the respective asset on trading day $t$. Sentiment indicators were aggregated by calendar date and then merged with the corresponding trading-day return series.

Because equity returns are only observed on trading days, non-trading days were excluded from the market return series. Reddit entries published on non-trading days were retained in the sentiment aggregation but aligned to the next available trading-day structure during time-series merging. This preserves the chronological ordering of information while ensuring that sentiment indicators and return observations are evaluated on a common market calendar.

Within individual calendar days, no exchange-time cutoff or timezone conversion was applied. The date was extracted directly from the available Reddit timestamp, and entries were aggregated over complete calendar days. Consequently, entries published after the U.S. market close remained assigned to their original calendar date rather than being transferred to the following trading day.

For the regression and classification analyses, the calendar-day sentiment indicator observed at day $t$ is related to return $R_{t+1}$. This design evaluates a next-day association, but it should not be interpreted as a strict real-time forecast based exclusively on information available before the beginning of the corresponding close-to-close return interval.

\subsection{Evaluation Design}

The empirical evaluation is designed to assess whether sentiment indicators exhibit market-relevant temporal structure and whether they contain information about subsequent return direction. All analyses are conducted separately for each asset and sentiment indicator.

First, lead/lag correlation analysis is used to examine temporal relationships between sentiment signals and stock returns. For each sentiment indicator $X_t$, correlations are computed over a symmetric window of $\pm 5$ trading days:

\begin{equation}
\rho_k = \operatorname{corr}(X_t, R_{t+k}), \quad k \in \{-5,\ldots,5\}.
\end{equation}

Positive values of $k$ indicate that sentiment is related to future returns and therefore test for potential predictive structure. Negative values of $k$ indicate that sentiment is related to past returns and therefore capture potentially reactive discourse patterns following market movements.

Second, Ordinary Least Squares (OLS) regression is used to evaluate whether lagged sentiment indicators explain next-day returns:

\begin{equation}
R_{t+1} = \alpha + \beta X_t + \epsilon_t,
\end{equation}

where $X_t$ denotes the respective sentiment indicator observed on day $t$, $R_{t+1}$ denotes the return on the following trading day, and $\epsilon_t$ denotes the error term. The regression is interpreted as an exploratory test of linear association rather than as a causal model.

Third, ROC-AUC is used to evaluate whether sentiment indicators contain information about the direction of next-day returns. For this purpose, the return direction is binarized as:

\begin{equation}
Y_{t+1} =
\begin{cases}
1, & \text{if } R_{t+1} > 0,\\
0, & \text{otherwise.}
\end{cases}
\end{equation}

The sentiment indicator $X_t$ is then used as a continuous score for classifying the binary direction variable $Y_{t+1}$. A ROC-AUC value of 0.5 indicates random directional classification, whereas values above 0.5 indicate directional information beyond random guessing.

The reported OLS p-values are based on conventional standard errors and are not adjusted for heteroskedasticity, serial dependence, or multiple testing. The lead/lag correlations are interpreted descriptively and are reported without formal confidence bands.

\subsection{Tail-Risk Early-Warning Framework}

To evaluate whether the LMI can serve as an early-warning indicator for extreme market regimes, a quantile-based backtesting framework is implemented. The framework tests whether unusually high discourse intensity is associated with an increased probability of extreme positive return events on the following trading day.

For the tail-risk evaluation, daily logarithmic returns are computed as:

\begin{equation}
r_t = \ln\left(\frac{P_t}{P_{t-1}}\right),
\end{equation}

where $P_t$ denotes the closing price on trading day $t$. Tail-risk events are defined as observations whose logarithmic return exceeds the rolling 95th percentile of the asset-specific historical return distribution over the preceding 30 trading days. Formally, the tail-risk indicator is defined as:

\begin{equation}
Tail_t =
\begin{cases}
1, & \text{if } r_t \geq Q_{0.95}(r_{t-30:t-1}),\\
0, & \text{otherwise.}
\end{cases}
\end{equation}

The rolling-window design ensures that the threshold is estimated using only historically available observations and therefore avoids look-ahead bias. It also adapts the threshold to time-varying volatility regimes.

Analogously, an LMI-based warning signal is generated using the rolling 95th percentile of the LMI distribution over the same 30-trading-day lookback window:

\begin{equation}
Signal_t =
\begin{cases}
1, & \text{if } LMI_t \geq Q_{0.95}(LMI_{t-30:t-1}),\\
0, & \text{otherwise.}
\end{cases}
\end{equation}

A signal generated on day $t$ is evaluated against the occurrence of a tail-risk event on the following trading day $t+1$. A true positive is recorded if $Signal_t = 1$ and $Tail_{t+1} = 1$. Because the underlying sentiment indicators are aggregated over complete calendar days without an exchange-time cutoff, this framework should be interpreted as an exploratory next-day early-warning backtest rather than as a strict real-time ex-ante forecasting design.

The classification performance of the early-warning framework is evaluated using precision and recall. Precision measures the share of emitted warning signals that are followed by a tail-risk event, whereas recall measures the share of actual tail-risk events that were preceded by a warning signal. The LMI is therefore not interpreted as a deterministic forecasting model, but as an exploratory indicator of elevated narrative intensity.

\section{Results}
\label{sec:results}

\subsection{Lexicon-Based Baseline Results}

Table~\ref{tab:vader_results} reports the regression and classification metrics for the VADER-based sentiment indicators. Across all three assets, the p-values of the OLS regressions remain well above the conventional significance level of $\alpha = 0.05$. The corresponding $R^2$ values are close to zero, indicating that the VADER-based sentiment indicators provide little explanatory power for next-day returns.

The ROC-AUC values are also close to random classification performance. For GME and AMC, the values remain around or below 0.5. NOK shows slightly higher ROC-AUC values, particularly for the weighted sentiment indicator, but the overall performance remains weak and does not indicate a robust directional signal.

\begin{table}[htbp]
\centering
\caption{VADER-Based Evaluation Metrics}
\label{tab:vader_results}
\begin{tabular}{llrrr}
\toprule
Asset & Indicator & $p$ & $R^2$ & AUC \\
\midrule
GME & Pure & 0.9073 & 0.0000 & 0.4832 \\
GME & Weighted & 0.8280 & 0.0001 & 0.4790 \\
AMC & Pure & 0.8168 & 0.0002 & 0.4776 \\
AMC & Weighted & 0.8630 & 0.0001 & 0.5100 \\
NOK & Pure & 0.6575 & 0.0021 & 0.5244 \\
NOK & Weighted & 0.5866 & 0.0032 & 0.5656 \\
\bottomrule
\end{tabular}
\end{table}

\subsection{LLM-Based Results}

Table~\ref{tab:llm_results} reports the regression and classification metrics for the LLM-based sentiment indicators. Compared to the VADER baseline, the LLM-derived indicators show somewhat stronger statistical structure, particularly for the composite LMI.

For the pure and score-weighted bullishness indicators, the OLS p-values remain above the conventional significance level of $\alpha = 0.05$ across all assets, and the corresponding $R^2$ values remain close to zero. This suggests that one-dimensional bullishness alone provides limited explanatory power for next-day returns.

The composite LMI yields more differentiated results. For AMC, the LMI exhibits a statistically significant negative association with next-day returns under conventional OLS inference ($\beta = -0.000041$, $p = 0.000005$), with an $R^2$ value of 0.0831. Thus, higher LMI values are associated with lower AMC returns on the following trading day in this specification. For GME and NOK, the LMI does not reach conventional significance levels, but its $R^2$ values remain higher than those of the corresponding pure and score-weighted bullishness indicators.

The ROC-AUC values indicate limited directional classification performance overall. Most values are close to random classification, while the LMI remains slightly above 0.5 for all three assets. These findings suggest that the LLM-based indicators provide a richer multidimensional representation and exhibit stronger asset-specific statistical structure than the lexicon-based baseline, but that this does not translate into consistently strong directional forecasting performance.

\begin{table}[htbp]
\centering
\caption{LLM-Based Evaluation Metrics}
\label{tab:llm_results}
\setlength{\tabcolsep}{3pt}
\begin{tabular}{@{}llccc@{}}
\toprule
Asset & Indicator & $p$ & $R^2$ & AUC \\
\midrule
GME & Bullishness & 0.6252 & 0.0007 & 0.5153 \\
GME & Weighted bullishness & 0.4746 & 0.0015 & 0.5238 \\
GME & LMI & 0.1313 & 0.0065 & 0.5273 \\
AMC & Bullishness & 0.9701 & 0.0000 & 0.4602 \\
AMC & Weighted bullishness & 0.5558 & 0.0014 & 0.4705 \\
AMC & LMI & $<0.0001$ & 0.0831 & 0.5176 \\
NOK & Bullishness & 0.8058 & 0.0006 & 0.4963 \\
NOK & Weighted bullishness & 0.6725 & 0.0016 & 0.5016 \\
NOK & LMI & 0.3330 & 0.0086 & 0.5058 \\
\bottomrule
\end{tabular}
\end{table}

\subsection{Lead/Lag Correlation Results}

The lead/lag correlation analysis provides additional insight into the temporal structure of the sentiment-return relationship. For the VADER-based indicators, correlations remain close to zero across the examined window of $\pm 5$ trading days. This supports the regression and ROC-AUC results reported in Table~\ref{tab:vader_results}, indicating that the lexicon-based sentiment signals do not exhibit a stable predictive or reactive relationship with returns in the analyzed setting.

\begin{figure*}[t]
\centering
\includegraphics[width=0.95\textwidth]{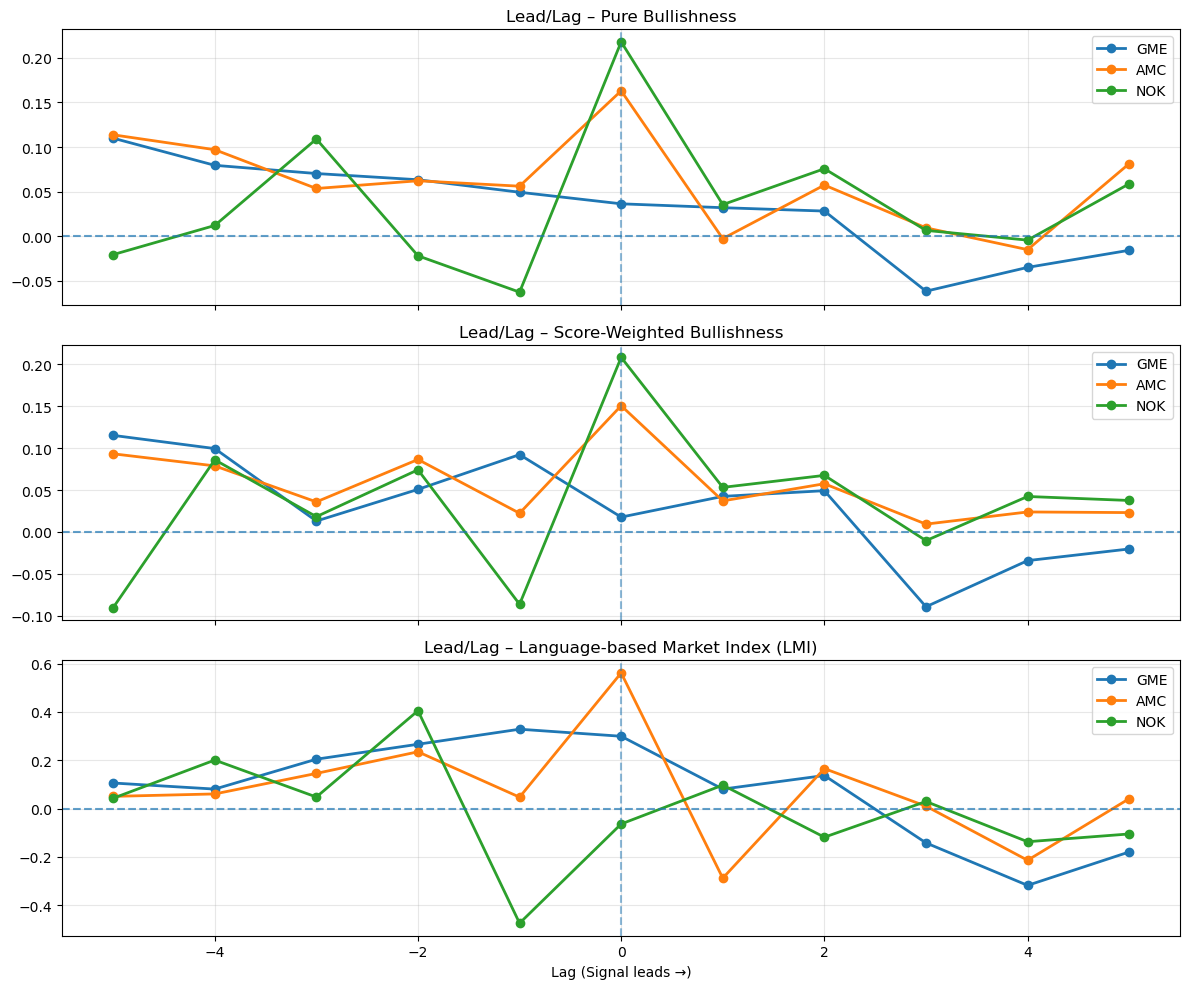}
\caption{Lead/lag correlations between LLM-derived discourse indicators and daily stock returns for GME, AMC, and NOK. Positive lags indicate that sentiment signals lead subsequent returns, whereas negative lags indicate reactive structures following prior market movements. The bottom panel reports the composite Language-based Market Index (LMI).}
\label{fig:leadlag_llm}
\end{figure*}

The LLM-based indicators show a more differentiated pattern. In particular, the composite LMI exhibits more pronounced correlation structures than the one-dimensional bullishness indicators. As shown in Figure~\ref{fig:leadlag_llm}, this pattern is most visible for AMC, where the LMI displays a pronounced contemporaneous correlation with returns followed by a reversal at the next lag. These findings suggest that multidimensional LLM-derived indicators capture stronger discourse-market interaction patterns than VADER-based sentiment, although the observed structures remain asset-dependent.

Overall, the lead/lag results indicate that social media discourse contains temporally structured information around extreme retail-driven market episodes. However, the evidence does not support a uniformly stable predictive relationship across assets. Instead, the results point to heterogeneous and partly reactive dynamics, consistent with the broader finding that richer linguistic representations do not necessarily translate into robust forecasting performance.

\subsection{Tail-Risk Early-Warning Results}

Table~\ref{tab:tail_results} reports the backtesting results of the LMI-based early-warning framework. In Table~\ref{tab:tail_results}, Obs. denotes valid observation days, Events denotes identified tail-risk events, Rate denotes the effective tail-event frequency, TP denotes true positives, Prec. denotes precision, and Rec. denotes recall. The effective tail rate serves as a baseline because a randomly emitted warning signal would be expected to achieve a precision approximately equal to the frequency of tail-risk events in the evaluation sample.

The results differ substantially across assets. For AMC, the LMI signal achieves a precision of 20.00\%, compared to an effective tail rate of 8.33\%. This indicates that warning signals are followed by tail-risk events more often than would be expected from the unconditional event frequency. However, recall remains limited at 24.24\%, meaning that only a minority of tail-risk events are preceded by a warning signal.

For GME, the early-warning performance is weak. Precision reaches 9.38\%, only slightly above the effective tail rate of 7.60\%, and recall remains low at 9.68\%. This suggests that the LMI provides little reliable early-warning structure for GME in the evaluated setting.

For NOK, the signal achieves the highest recall at 52.17\%, indicating that more than half of the identified tail-risk events are preceded by an LMI warning signal. However, precision remains low at 8.82\%, only moderately above the effective tail rate of 6.32\%. This implies that the signal captures a relatively large share of tail events but also produces many warnings that are not followed by such events.

Overall, the results suggest that the LMI can contain tail-risk-related information in selected cases, but its performance is asset-dependent and not sufficiently stable to support interpretation as a robust standalone forecasting model.

\begin{table}[htbp]
\centering
\caption{LMI-Based Tail-Risk Backtesting Results}
\label{tab:tail_results}
\setlength{\tabcolsep}{3pt}
\begin{tabular}{@{}lrrrrrr@{}}
\toprule
Asset & Obs. & Events & Rate & TP & Prec. & Rec. \\
\midrule
AMC & 396 & 33 & 8.33\% & 8 & 20.00\% & 24.24\% \\
GME & 408 & 31 & 7.60\% & 3 & 9.38\% & 9.68\% \\
NOK & 364 & 23 & 6.32\% & 12 & 8.82\% & 52.17\% \\
\bottomrule
\end{tabular}
\end{table}

\section{Discussion}
\label{sec:discussion}

\subsection{Interpretation of Findings}

The results indicate that LLM-based sentiment representations provide a richer multidimensional structure and exhibit stronger asset-specific statistical associations than the lexicon-based VADER baseline. Across the evaluated assets, VADER-based indicators show little explanatory or directional classification power, with regression results close to zero and ROC-AUC values near random performance. This suggests that one-dimensional lexicon-based polarity measures provide limited market-relevant signal in the evaluated setting.

The LLM-based indicators provide a more differentiated picture. In particular, the composite LMI exhibits stronger statistical structure than the pure and score-weighted bullishness indicators. The strongest result is observed for AMC, where the LMI shows a statistically significant negative association with next-day returns and a higher explained variance than any VADER-based specification. This supports the view that multidimensional language representations can capture market-relevant discourse features that are not reflected in simple polarity scores.

However, the results also show that richer linguistic representation does not automatically imply robust forecasting performance. While the LMI improves over VADER in several respects, the predictive signal remains asset-dependent and generally weak in directional classification. ROC-AUC values remain close to random classification performance, and the lead/lag results suggest that part of the observed relationship may be contemporaneous or reactive rather than strictly predictive.

These findings provide a nuanced answer to the research questions. Regarding RQ1, the LLM-based framework provides a multidimensional representation that separates emotional polarity, market intent, topical relevance, and sarcasm, whereas VADER produces a one-dimensional polarity score. Regarding RQ2, the lead/lag, regression, and classification results indicate heterogeneous and partly reactive relationships rather than a stable cross-asset predictive pattern. The LLM-based indicators exhibit stronger statistical structure in selected cases, but their directional classification performance remains weak overall. Regarding RQ3, the LMI provides limited early-warning information for extreme positive return events, most notably for AMC, but its precision and recall are not sufficiently stable to support interpretation as a robust standalone forecasting model.

\subsection{Limitations}

Several limitations should be considered when interpreting the results. First, the analysis is restricted to Reddit discourse from r/WallStreetBets and to three meme stocks: GME, AMC, and NOK. The findings therefore reflect a specific retail-driven market environment and cannot be assumed to generalize to broader equity markets or other information channels.

Second, the dataset is strongly imbalanced across assets, data sources, and entry types, with Figshare comments accounting for the majority of observations. In addition, truncating Reddit entries to the first 400 characters may omit relevant context from longer submissions and comments.

Third, the study does not use manually validated ground-truth labels for sentiment, bullishness, sarcasm, or relevance. The LLM-derived indicators are evaluated indirectly through their relationship with market data rather than through semantic annotation. Moreover, the comparison with VADER is not fully construct-equivalent, since VADER produces a general polarity score whereas the LLM-based framework also captures bullishness and combines several semantic and social dimensions. In addition, the engagement-weighting schemes are not identical across indicators: the weighted VADER and bullishness indices retain raw negative Reddit scores, whereas non-positive scores receive zero weight in the LMI. This asymmetry may affect the comparability of the resulting indicators. The empirical comparison is also limited to VADER and a single LLM configuration. Accordingly, the findings should not be generalized to all lexicon-based, transformer-based, or LLM-based methods.

Fourth, the statistical results should be interpreted as exploratory associations rather than causal evidence. Multiple asset--indicator specifications and lead/lag relationships were examined, and the analysis does not explicitly account for multiple testing, heteroskedasticity, or serial dependence in daily returns. In addition, Reddit entries were aggregated over complete calendar days without an exchange-time cutoff or timezone conversion. The next-day analyses may therefore include entries published after the beginning of the corresponding close-to-close return interval and should not be interpreted as strict real-time forecasts.

Finally, Reddit scores provide only an approximate measure of social visibility and engagement, and the LLM-based pipeline depends on external model infrastructure. This introduces limitations regarding model versioning, reproducibility, and transfer to real-time applications.

\subsection{Implications}

The findings suggest that LLM-based sentiment analysis can be useful as a research tool for extracting structured discourse indicators from large-scale financial social media data. In particular, multidimensional representations can distinguish discourse dimensions such as emotional polarity, trading intent, sarcasm, and relevance that are not represented by simple polarity measures.

For financial risk monitoring, the LMI should be interpreted as a narrative intensity indicator rather than as a deterministic trading or forecasting signal. Elevated LMI values may indicate periods of intensified discourse that coincide with or precede extreme market movements, but the observed precision and recall values are not stable enough for standalone operational use.

Overall, the results support the use of LLM-based discourse indicators as complementary inputs for market monitoring and risk analysis. Their main value lies not in replacing quantitative market models, but in enriching them with information about narrative intensity and retail-driven attention dynamics.

\section{Conclusion}
\label{sec:conclusion}

This paper examined whether Large Language Model-based sentiment representations provide more informative market signals than a lexicon-based baseline in the context of social media-driven equity markets. Using Reddit data from r/WallStreetBets and focusing on GME, AMC, and NOK, we compared VADER-based sentiment indicators with LLM-derived multidimensional indicators, including sentiment, bullishness, sarcasm, relevance, and the composite Language-based Market Index (LMI).

The results show that VADER-based sentiment indicators provide little explanatory or directional classification power in the evaluated setting. In contrast, the LLM-based indicators, particularly the LMI, provide a richer multidimensional representation and exhibit stronger statistical associations in selected cases. The most pronounced result is observed for AMC, where the LMI shows a significant negative relationship with next-day returns and improved explanatory power compared to the lexicon-based baseline.

However, the findings also demonstrate that richer linguistic representation does not automatically translate into robust market predictability. Directional classification performance remains weak overall, and the observed relationships vary substantially across assets. The LMI-based early-warning framework provides evidence of tail-risk-related information in selected cases, but its performance is not sufficiently stable to support interpretation as a standalone forecasting system.

Overall, the study suggests that LLM-based sentiment analysis can support the extraction of structured discourse signals from financial social media data. Its main value lies in capturing narrative intensity, trading intent, and context-sensitive communication patterns that are difficult to represent with simple polarity-based sentiment models. For financial risk monitoring, such indicators should therefore be interpreted as complementary narrative signals rather than deterministic trading or prediction tools.

Future research could extend the analysis to additional assets, longer observation windows, and other social media platforms. Further work should also compare different LLM architectures, incorporate manually validated sentiment labels, and evaluate alternative engagement-weighting schemes. Finally, real-time implementations would require robust model-version control, low-latency inference, and continuous validation under changing market and discourse conditions.

\appendix
\section{LLM Classification Prompt}
\label{app:prompt}

The following ticker-parameterized system prompt was applied consistently across all inference batches:

\begin{Verbatim}[
  fontsize=\footnotesize,
  breaklines=true,
  breakanywhere=true,
  breaksymbolleft={}
]
Analyze Reddit posts from r/wallstreetbets related to the {TICKER} stock.

STRICT RULES:
- Each input has an immutable `id`.
- Copy the `id` EXACTLY as given in the input.
- Do not modify, omit, regenerate, or reorder ids.
- Output must contain exactly one result per input, in the same order.

WSB language rules:
- Profanity, ALL CAPS, insults, memes, exaggeration -> often bullish hype.
- "I like the stock", YOLO, apes, diamond hands -> bullish.
- Negative wording can be bullish if ironic or hype-driven.
- Sarcasm is common.
- Buying, holding, FOMO, mocking bears -> bullish.
- Neutral/bot posts -> s=0, b=0, sc=0, r=0.

Output ONLY minified JSON array:
[id, s, b, sc, r]

Definitions:
id = id given in the input (unchanged)
s = sentiment (-1..1)
b = bullishness (-1..1)
sc = sarcasm (0..1)
r = relevance to {TICKER} (0..1)

No text. No explanations.
\end{Verbatim}

\bibliographystyle{abbrvurl}
\bibliography{references}

\end{document}